\documentclass[11pt]{article}
\usepackage[margin=1in]{geometry}
\usepackage{setspace}
\usepackage{amsmath,amssymb}
\usepackage{graphicx}
\usepackage{booktabs}
\usepackage{array}
\usepackage{framed}
\usepackage[numbers,sort&compress]{natbib}
\usepackage{url}
\usepackage[hidelinks]{hyperref}
\usepackage{microtype}

\providecommand{\dropcap}[1]{#1}

\begin{document}
\begin{center}
{\LARGE\bfseries Topological summaries of fingerprint ridge patterns carry identity information\par}
\vspace{0.8em}
{\large Chad M. Topaz$^{a,b,c,\ast}$\quad Niny Arcila-Maya$^{d}$\quad Elizabeth Munch$^{e}$\\[3pt]
Zofia Stanley$^{b}$\quad Lori Ziegelmeier$^{f}$\par}
\vspace{0.6em}
{\footnotesize
$^{a}$Williams College, Williamstown, MA 01267\\
$^{b}$QSIDE Institute, Williamstown, MA 01267\\
$^{c}$University of Colorado Boulder, Boulder, CO 80309\\
$^{d}$San Francisco State University, San Francisco, CA 94132\\
$^{e}$Michigan State University, East Lansing, MI 48824\\
$^{f}$Macalester College, Saint Paul, MN 55105\\[4pt]
$^{\ast}$To whom correspondence should be addressed. E-mail: cmt6@williams.edu\par}
\end{center}
\vspace{0.4em}

\begin{abstract}
Fingerprints are the most widely deployed biometric. Verifying whether two impressions come from the same finger typically relies on minutiae, small landmarks such as skin ridge endings and bifurcations. These landmarks are extracted through a multi-stage pipeline of image enhancement, skeletonization, minutiae detection, and alignment. We investigate an alternative: using topological data analysis to represent the full pattern of skin ridges and valleys directly, bypassing minutiae detection and the downstream matching pipeline. We apply persistent homology, a topological tool that tracks how loops in the ridge pattern form and fill in across spatial scales, producing multi-scale summaries of ridge geometry. We develop and compare a range of verification methods on a standard benchmark dataset, FVC2000 DB1. Even the simplest topological summaries, with no trained parameters, substantially outperform geometry-only baselines. A trained method achieves an AUC of 0.91, while an optimal-transport method excels at the strictest false-accept thresholds, suggesting they capture different aspects of the ridge pattern. Fusing these two approaches yields the best performance at every low false-accept threshold we examine. Our results establish that these topological summaries capture substantial fingerprint identity information, far more effective for verification than raw pixel-level geometry. Because the entire pipeline is openly specified, it offers a transparent complement to minutiae-based systems, and we provide a modular framework for constructing, evaluating, and combining topological verification methods.
\end{abstract}

\begin{framed}\noindent\textbf{Significance Statement.}\quad Fingerprint matching is central to identification systems used by law enforcement and consumer devices worldwide. Most matching systems depend on minutiae, tiny ridge landmarks extracted through a long, partly proprietary processing pipeline whose reliability has drawn scientific scrutiny. We ask how much identity information remains when those landmarks are discarded and a fingerprint is summarized by the overall shape of its ridge pattern, using topological data analysis. Built from openly specified rather than proprietary steps, these shape summaries substantially outperform direct pixel comparison on a standard benchmark, and combining two complementary summaries is most effective at the stringent error rates security applications demand. The findings point toward hybrid systems that strengthen conventional matchers with additional ridge-shape features.\end{framed}

\section{Introduction}

\dropcap{F}ingerprints are the oldest and most widely deployed biometric~\cite{maltoni2022handbook}. Law enforcement agencies worldwide maintain databases containing tens of millions of records, commercial devices from smartphones to border control kiosks authenticate users by fingerprints daily, and latent prints, the often-invisible impressions left unintentionally at crime scenes, continue to play a central role in criminal investigations~\cite{maltoni2022handbook}. The societal stakes of fingerprint matching are correspondingly high: a false accept can grant unauthorized access or implicate an innocent person, while a false reject can deny a legitimate user or allow a suspect to evade identification. This work concerns fingerprint \emph{verification}, deciding whether two impressions come from the same finger, rather than searching a database to identify an unknown print.

Most large-scale fingerprint systems extract \emph{minutiae}, discrete landmarks such as ridge endings and bifurcations, from a processed fingerprint image, and then compare minutiae configurations across images using spatial matching algorithms~\cite{maltoni2022handbook,yager2004fingerprint,valdes2019review}. This pipeline involves a cascade of choices: image enhancement filters, binarization thresholds, skeletonization heuristics, minutiae detection rules, and alignment procedures. The downstream steps in particular introduce many parameters and design decisions, and the specific algorithms used by commercial systems remain largely proprietary~\cite{daluz2018fundamentals}.

The reliability of this paradigm has come under sustained scientific scrutiny, particularly in the forensic setting of latent print examination. A landmark 2009 report by the National Research Council concluded that, with the exception of nuclear DNA analysis, no forensic method, including fingerprint examination, had been rigorously shown to have the capacity to establish a connection between evidence and a specific individual~\cite{NRC2009}. A subsequent 2016 report by the President's Council of Advisors on Science and Technology found that latent fingerprint analysis, while foundationally valid, exhibited false positive rates substantially higher than commonly assumed, reaching as high as one erroneous identification in 18 cases~\cite{PCAST2016}. These assessments concern latent examination rather than automated verification of known prints, but they highlight a broader point: the minutiae-based paradigm, despite decades of operational use, rests on an empirical foundation that has not been fully characterized.

Against this backdrop, we explore alternative mathematical representations of fingerprint images: representations whose properties can be derived analytically rather than only validated empirically, as is the case for minutiae, and that reduce dependence on the multi-stage feature-extraction cascade. Topological data analysis (TDA) offers such a framework. TDA studies the ``shape'' of data across multiple scales, and a central tool, persistent homology, encodes features such as connected components and loops in a summary that is stable with respect to perturbations of the input~\cite{otter2017roadmap,wasserman2018topological,carlsson2020topological}. These properties suit fingerprint images, whose ridge-and-valley patterns naturally give rise to multi-scale topological structure. Topological loops, pockets of valley enclosed by ridges and distinct from the ``loop'' pattern type, form in inter-ridge spaces, the arrangement of bifurcations and endings shapes which loops form and how long they persist as ridges are progressively thickened, and the global pattern type reflects the topological organization of the ridge flow. Despite this fit, topological methods have rarely been applied to fingerprints, and the few prior efforts still rely on conventional minutiae detection. Earlier work applied persistent homology to point clouds of minutiae for fingerprint classification into arch, loop, and whorl types~\cite{giansiracusa2019persistent}, and a recent short conference paper computed persistent homology of extracted minutiae points and compared the resulting diagrams for matching~\cite{devkar2025topological}. In both, minutiae are detected first and topology serves only as a downstream classifier or distance, so persistent homology supplements the standard minutiae pipeline rather than replacing it as the primary representation.

We investigate a deliberately constrained question: \emph{how much fingerprint identity information can be captured by topological representations alone, without recourse to minutiae extraction, skeletonization, or explicit alignment?} Starting from standard enhanced binary ridge/valley images, we compute two Euclidean distance transforms, which give each pixel's distance to the nearest ridge or to the nearest valley, and apply sublevel set persistent homology to these fields. We use these topological summaries, together with distance-transform geometry baselines, to construct similarity scores for seven verification methods spanning three families, evaluated on FVC2000 DB1~\cite{maio2002fvc}, a standard benchmark of 100 identities with 8 impressions each. Geometry-only baselines, which compare downsampled distance transforms directly at two resolutions, achieve an area under the ROC curve (AUC) near 0.71 and establish a floor for non-topological template matching. Two global topological methods, one based on Betti curves and one on persistence images, require no learning and achieve AUC near 0.85. A logistic-regression model trained on symmetric pairwise topological features achieves AUC 0.906 with an equal error rate (EER) of 17.4\%, while an optimal-transport method that matches local topological neighborhoods between fingerprints excels at low false accept rates, reaching a true accept rate (TAR) of 11.2\% at a false accept rate (FAR) of $10^{-3}$ versus 4.9\% for the learned model at the same threshold. A simple fusion of the two achieves the best true accept rate at all reported low-FAR operating points. All experiments use identity-disjoint 5-fold cross-validation with strict no-leakage protocols: every fold-specific threshold, score-normalization parameter, learned model coefficient, and fusion weight is estimated from training data only.

These results establish that persistent homology of distance transform fields encodes substantial fingerprint identity information, far more effective for verification than raw pixel-level geometry, and provide a rigorous, reproducible framework for extracting and evaluating it. They do not suggest that topology should replace minutiae-based systems, which achieve substantially lower error rates by exploiting the very features, minutiae and explicit alignment, that we deliberately exclude. Because our topological representations summarize the ridge pattern differently from minutiae templates, they are a natural complement to conventional features in hybrid systems.

\section{Topological Representation of Fingerprints}

\begin{figure}[!ht]
\centering
\includegraphics[width=\linewidth]{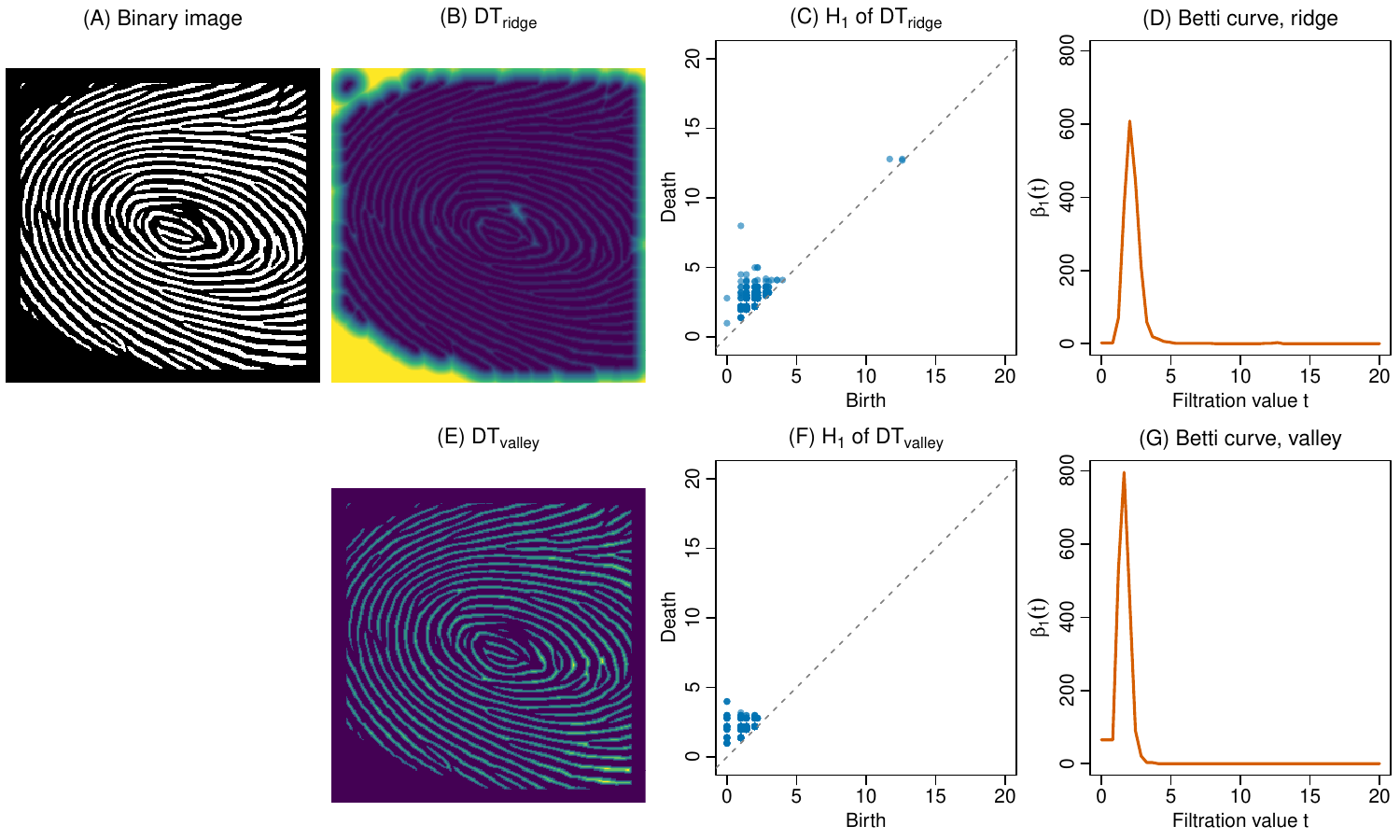}
\caption{The topological representation pipeline applied to an example fingerprint from FVC2000 DB1. The top row shows the ridge channel; the bottom row, the valley channel. (\emph{A})~Binary ridge image after segmentation and polarity normalization, the common input for both channels. (\emph{B} and \emph{E})~Ridge and valley distance transforms $\mathrm{DT}_\mathrm{ridge}$ and $\mathrm{DT}_\mathrm{valley}$, giving each pixel's Euclidean distance to the nearest ridge or valley pixel respectively; lighter values indicate greater distance. (\emph{C} and \emph{F})~$H_1$ persistence diagrams computed from the sublevel set filtrations of $\mathrm{DT}_\mathrm{ridge}$ and $\mathrm{DT}_\mathrm{valley}$ at the $150 \times 150$ working resolution; each point $(b,d)$ represents a loop that appears at filtration value $b$ and is filled in at value $d$. (\emph{D} and \emph{G})~Corresponding Betti curves $\beta_1(t)$, counting the number of loops alive at each filtration value. The ridge and valley channels produce visibly different persistence diagrams and Betti curves, reflecting their complementary geometric content.}
\label{fig:pipeline}
\end{figure}

Our pipeline converts each binary fingerprint image into topological feature vectors through three stages: distance transform computation, persistent homology, and vectorization. Figure~\ref{fig:pipeline} illustrates the full sequence on an example fingerprint.

\subsection{Distance transforms}

The input images are binary ridge/valley segmentations produced by a standard enhancement algorithm~\cite{hong1998fingerprint}; Fig.~\ref{fig:pipeline}\emph{A} shows an example. We deliberately omit four standard fingerprint processing steps from our topological pipeline: ridge thinning or skeletonization, estimation of orientation fields, minutiae detection, and Gabor filtering or other ridge enhancement beyond the upstream preprocessing. We acknowledge that the upstream enhancement algorithm~\cite{hong1998fingerprint} internally uses orientation estimation and Gabor filtering to produce the binary segmentations that serve as our input; our claim is that the \emph{topological representation itself} requires none of these operations, not that orientation information is entirely absent from the pipeline. Our goal is to evaluate how much information can be extracted from topology alone, starting from standard enhanced binary ridge patterns.

From each binary image, in which ridge pixels form the minority class, we compute two Euclidean distance transform fields. The \emph{ridge distance transform} assigns to each pixel $x$ its Euclidean distance to the nearest ridge pixel,
\begin{equation}
\label{eq:dt_ridge}
\mathrm{DT}_{\mathrm{ridge}}(x) = \min_{y \in \mathrm{ridges}} \|x - y\|_2,
\end{equation}
and the \emph{valley distance transform} assigns to each pixel its distance to the nearest valley pixel,
\begin{equation}
\label{eq:dt_valley}
\mathrm{DT}_{\mathrm{valley}}(x) = \min_{y \in \mathrm{valleys}} \|x - y\|_2.
\end{equation}
These two fields are complementary. The ridge distance transform is zero on ridge pixels and increases as one moves away from ridges into valley regions, reaching local maxima at points equidistant from neighboring ridges. The valley distance transform has the opposite behavior: zero on valleys, increasing toward ridge centers. Together, they encode the geometry of the ridge-valley pattern without any explicit feature detection. The distance transform converts a discrete binary pattern into a real-valued scalar field, the required input for sublevel set persistent homology. It encodes ridge width and spacing in a natural way. It is also simple and parameter-free, uniquely determined by the binary image. We carry both fields through the topological pipeline (Fig.~\ref{fig:pipeline}).

\subsection{Sublevel set persistent homology}

Persistent homology tracks topological features across a range of scales. Given a function $f$ defined on a domain, the sublevel set at scale $t$ is
\begin{equation}
\label{eq:sublevel}
L_t(f) = \{x : f(x) \leq t\}.
\end{equation}
As $t$ increases from the minimum to the maximum value of $f$, the sublevel set grows monotonically, forming a nested sequence of spaces. For a function on a pixel grid, this filtration is realized on a \emph{cubical complex} whose 2-dimensional cells are pixels, edges between adjacent pixels, and vertices at pixel corners. Each pixel takes its function value, and each edge and vertex takes the minimum value over the pixels incident to it. Homology quantifies the topology of each space in the filtration: $\beta_0$ counts connected components and $\beta_1$ counts loops, with corresponding homology groups $H_0$ and $H_1$. Each feature has a \emph{birth} scale $b$, the parameter value at which it first appears, and a \emph{death} scale $d$, the value at which it disappears. The collection of all birth--death pairs forms the \emph{persistence diagram}, and the \emph{persistence} $d - b$ is the range of scales over which a feature persists. A key property is \emph{stability}: small perturbations of the input function produce small perturbations of the persistence diagram~\cite{otter2017roadmap}. For fingerprint verification, minor imaging variations that produce small pointwise changes in the distance transform produce correspondingly small changes in the persistence diagram. The guarantee does not extend to large geometric deformations such as elastic stretching of the skin; robustness to such deformations is an empirical property of our methods, not a theoretical one. \emph{SI Appendix}, section~\ref{si:background} gives a fuller treatment, including the behavior of rotation invariance on discrete grids.

\begin{figure}[!ht]
\centering
\includegraphics[width=\linewidth]{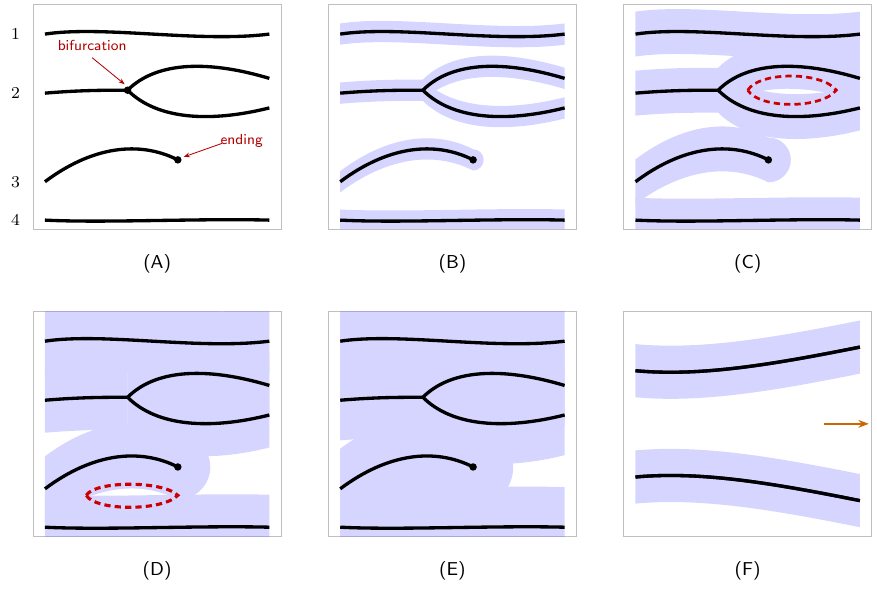}
\caption{Schematic of the sublevel set filtration of the ridge distance transform.
(\emph{A})~A stylized configuration of four ridges $1$--$4$: ridge~$2$ splits at a bifurcation into two branches that diverge and then partially reconverge, bounding a valley region between them; ridge~$3$ terminates at a ridge ending, leaving a wide gap to ridge~$4$.
(\emph{B}--\emph{E})~As the scale~$t$ increases, tubes of radius~$t$ expand around each ridge and progressively fill the image.
(\emph{B})~At small~$t$, all tubes are thin and separated; no valley regions are enclosed.
(\emph{C})~At moderate~$t$, the tubes from the two branches of ridge~$2$ merge to enclose a valley, giving birth to an $H_1$ feature (dashed red curve); the relatively narrow opening between the converging branches closes quickly, producing a low birth scale.
(\emph{D})~At larger~$t$, the $H_1$ feature from the two branches of ridge~$2$ has died, and the tube around ridge~$3$ reaches ridge~$4$, enclosing the valley between them and giving birth to a second $H_1$ feature; the wide gap left by the ridge ending delays this birth relative to the bifurcation loop.
(\emph{E})~At still larger~$t$, the enclosed valley region is completely filled in and the second $H_1$ feature dies; the death scale of each feature corresponds to the inradius of the enclosed region.
(\emph{F})~A valley region that is open to the image boundary (orange arrow) is never fully enclosed by the expanding tubes and therefore does not produce an $H_1$ feature.}
\label{fig:filtration}
\end{figure}

The sublevel sets of the ridge distance transform have an intuitive geometric interpretation, illustrated in Fig.~\ref{fig:filtration}. Because the distance transform records distance to the ridge set in pixels, the scale $t$ is a spatial quantity measured in pixels: the sublevel set at scale $t$ contains every pixel within distance $t$ of the ridges, a tube of radius $t$. At scale $t = 0$, the sublevel set consists of exactly the ridge pixels. As $t$ increases, the sublevel set expands to include pixels progressively farther from the ridges, forming a growing tube around the ridge set. Initially separate ridge components merge as their tubes meet, producing deaths of $H_0$ features. Loops then form in the inter-ridge spaces as the expanding tubes enclose valley regions, producing births of $H_1$ features; as $t$ increases further, these loops are filled in and die.

We focus on $H_1$ for fingerprint analysis. An $H_1$ feature is born when the expanding tube around the ridge set completely encloses a region of valley space, and it dies when that region is completely filled in. For a valley region with multiple openings to the exterior, the loop is born when the last opening closes, controlled by the widest remaining gap between the bounding ridges. The death scale is governed by the \emph{inradius} of the enclosed valley region, the maximum distance from any interior point to the bounding ridges, rather than by the valley's area. Minutiae affect this process through concrete geometric mechanisms. At a ridge bifurcation, a Y-shaped junction bounds a small valley region on three sides; the expanding tubes close off the region quickly, producing an $H_1$ feature with a low birth scale and short persistence (Fig.~\ref{fig:filtration}\emph{C}). At a ridge ending, one of the bounding ridge segments terminates, leaving a wide gap that the expanding tube must cross before the region is enclosed; this delays the birth of the corresponding loop (Fig.~\ref{fig:filtration}\emph{D}) or, if the gap is too wide, prevents it entirely. The $H_1$ persistence diagram thus encodes a multi-scale signature of how ridges are arranged, spaced, and branched, precisely the geometric features that vary between different fingers. Valley regions along the image boundary that remain connected to the exterior never produce $H_1$ features (Fig.~\ref{fig:filtration}\emph{F}), so the $H_1$ signal is driven primarily by interior ridge structure. We do not use $H_0$ features in our scoring methods: all connected components of the ridge set are born at $t = 0$ and merge rapidly, offering little discriminative variation across identities.

We compute persistent homology separately on the ridge and valley distance transforms, yielding two $H_1$ persistence diagrams per finger image. The two filtrations grow from different sets and are not related by a simple reparameterization; the valley-centric perspective is sensitive to ridge thickness and arrangement, complementing the ridge-centric view.

\subsection{From diagrams to feature vectors}

A persistence diagram is a multiset of points in the plane, and direct comparison between diagrams is computationally cumbersome. We therefore convert diagrams into fixed-length vectors using two complementary techniques. The \emph{Betti curve} counts how many features are alive at each filtration value,
\begin{equation}
\label{eq:betti}
\beta_1(t) = \#\{(b,d) \in D : b \leq t < d\},
\end{equation}
where $D$ denotes the $H_1$ persistence diagram. We evaluate the curve on a grid of 50 equally spaced values, and concatenating ridge and valley curves gives a 100-dimensional \emph{global Betti vector}, normalized by ridge area fraction so that the vector reflects the shape of the topological profile rather than the overall amount of ridge content. \emph{Persistence images}~\cite{adams2017persistence} provide an alternative vectorization: each diagram point is mapped to birth--persistence coordinates, weighted by a function that vanishes at zero persistence, convolved with a Gaussian kernel, and discretized on a regular grid to yield a vector capturing the joint distribution of birth scales and persistence values. With this weighting, persistence images inherit a stability guarantee analogous to that of the underlying diagram~\cite{adams2017persistence}; Betti curves do not enjoy the same guarantee in the bottleneck metric.

In addition to global summaries computed from the entire image, some of our methods use \emph{local} summaries that capture how topological structure varies across the fingerprint. Corresponding regions of two impressions from the same finger should look alike, so we sample 64 anchor points uniformly at random from ridge pixels, crop a circular patch of the distance transform fields around each anchor, mask the patch boundary to avoid spurious features, and compute local $H_1$ persistence diagrams and their Betti curves or persistence images. Patches near bifurcations, lakes, and tightly curved ridge arrangements produce rich local diagrams, while patches in regions of approximately parallel ridge flow often produce few or no features; this patch-to-patch variability is the signal exploited by the local methods below. Full extraction details and an illustration appear in \emph{SI Appendix}, section~\ref{si:rep} and Fig.~\ref{fig:patches}.

\section{Verification Methods}

We evaluate seven methods that produce a similarity score from an image pair, spanning three families: geometry-only baselines requiring no topological computation, global topological comparisons requiring no learning, and methods that incorporate learning or structured matching. Key parameter values for all methods appear in \emph{SI Appendix}, Table~\ref{tab:params}, with the complete specification in \emph{SI Appendix}, section~\ref{si:methods}.

\subsection{Geometry baselines}

Two geometry-only baselines establish a floor for what can be achieved without persistent homology. Both concatenate the ridge and valley distance transforms into a single vector, normalize by root-mean-square value, and score pairs by negative Euclidean distance. \emph{DT Downsampled $L^2$ ($50 \times 50$)} uses distance transforms downsampled to $50 \times 50$ grids; \emph{DT Downsampled $L^2$ ($150 \times 150$)} uses the same $150 \times 150$ resolution as all topological methods. The coarser baseline establishes a deliberately weak floor, while the matched-resolution baseline controls for input resolution, so that any advantage of topological methods cannot be attributed to a difference in the underlying input. Any topological method that fails to beat these baselines would contribute no useful information beyond what the distance transform images already contain.

\subsection{Global topological baselines}

Two methods test whether global topological summaries alone, with no learning, can discriminate fingerprints. \emph{Global Betti Curve $L^1$} scores a pair by the negative $L^1$ distance between global Betti vectors; the $L^1$ distance corresponds to the total discrepancy in feature counts integrated over the filtration, a natural measure for count-based summaries. \emph{Global Persistence Image $L^2$} scores a pair by the negative $L^2$ distance between concatenated, $L^2$-normalized ridge and valley persistence images. Both methods have no learned parameters and require no training data, providing a clean measurement of how much discriminative information is present in the topological summaries before any learning is applied.

\subsection{TopoLR: learned topological pair scoring}

\emph{TopoLR} learns which aspects of the topological representation are most discriminative for verification. Each image is represented by a 400-dimensional vector concatenating the global Betti vector (100 dimensions) with pooled local Betti statistics (300 dimensions: mean across all 64 anchors, mean across the 10 most topologically active anchors within each channel, and elementwise standard deviation, for the ridge and valley channels). Given two images with feature vectors $u, v \in \mathbb{R}^{400}$, we construct a symmetric pair representation invariant under interchange of the images: the elementwise absolute difference $|u - v|$, minimum $\min(u,v)$, maximum $\max(u,v)$, and product $u \odot v$, concatenated into a 1{,}600-dimensional pair feature vector. The absolute difference captures discrepancy, the min and max capture shared and extremal structure, and the product captures co-occurrence. We train an $L^2$-regularized logistic regression model on training pairs; the output score is the logistic model output. The choice of logistic regression is deliberate: with 1{,}600 features and roughly 200{,}000 training pairs per fold, overfitting is a concern with more flexible classifiers, and the learned weights remain interpretable. Training details appear in \emph{Materials and Methods}.

\subsection{LPOT: local persistence image optimal transport}

While TopoLR learns a global similarity function from pooled features, \emph{LPOT} explicitly matches local regions across two images using optimal transport, a framework for finding the cheapest way to match the elements of two collections. Corresponding regions of two impressions from the same finger should have similar local topology, but their spatial positions may shift due to differences in finger placement, pressure, and elastic skin deformation. For each image, we compute a 1{,}800-dimensional local persistence-image descriptor $d_i$ at each of the 64 anchors, concatenating $L^2$-normalized ridge and valley channels, and record each anchor's normalized spatial coordinates. The cost of matching anchor $i$ in image $A$ to anchor $j$ in image $B$ combines a \emph{descriptor cost} measuring topological dissimilarity,
\begin{equation}
\label{eq:desc_cost}
C^{\mathrm{desc}}_{ij} = \|d_i - d_j\|_2^2,
\end{equation}
where $d_i$ and $d_j$ are the descriptors, with a \emph{geometry cost} measuring spatial distance,
\begin{equation}
\label{eq:geo_cost}
C^{\mathrm{geo}}_{ij} = \|p_i - p_j\|_2^2,
\end{equation}
where $p_i$ and $p_j$ are the normalized positions, combined as
\begin{equation}
\label{eq:total_cost}
C_{ij} = C^{\mathrm{desc}}_{ij} + \beta \cdot C^{\mathrm{geo}}_{ij}
\end{equation}
with geometry weight $\beta = 6$. The geometry cost encourages matches that are not only topologically similar but also spatially consistent. Some anchors may correspond to noise, damaged regions, or areas outside the overlap between two impressions, so we use a \emph{dustbin mechanism} permitting partial matching: an augmented transport problem allows up to 20\% of the mass on each side to route to a virtual node at fixed cost, and at least 75\% of the real-anchor mass must still be matched between real anchors. We solve the entropically regularized problem with the Sinkhorn algorithm using a fixed iteration count, and the similarity score is the negative total transport cost. While LPOT does not perform explicit geometric alignment or registration, its geometry cost encourages spatially coherent matches and thus implicitly benefits from approximate positional consistency across impressions. This is a softer spatial constraint than classical alignment but is not alignment-free.

\subsection{Fusion}

TopoLR and LPOT have complementary strengths. We therefore evaluate a simple linear fusion. We normalize each method's scores by a robust z-score transformation, subtracting the median and dividing by the median absolute deviation (MAD) of the training impostor scores, and combine
\begin{equation}
\label{eq:fusion}
s_{\mathrm{fusion}} = w \cdot z_{\mathrm{TopoLR}} + (1-w) \cdot z_{\mathrm{LPOT}},
\end{equation}
with $w$ selected by grid search on training data to maximize TAR at FAR $= 10^{-3}$. With this simple design we show that the complementarity between TopoLR and LPOT can be exploited without complex model stacking. A single scalar parameter suffices.

\section{Results}

\begin{figure}[!ht]
\centering
\includegraphics[width=0.78\linewidth]{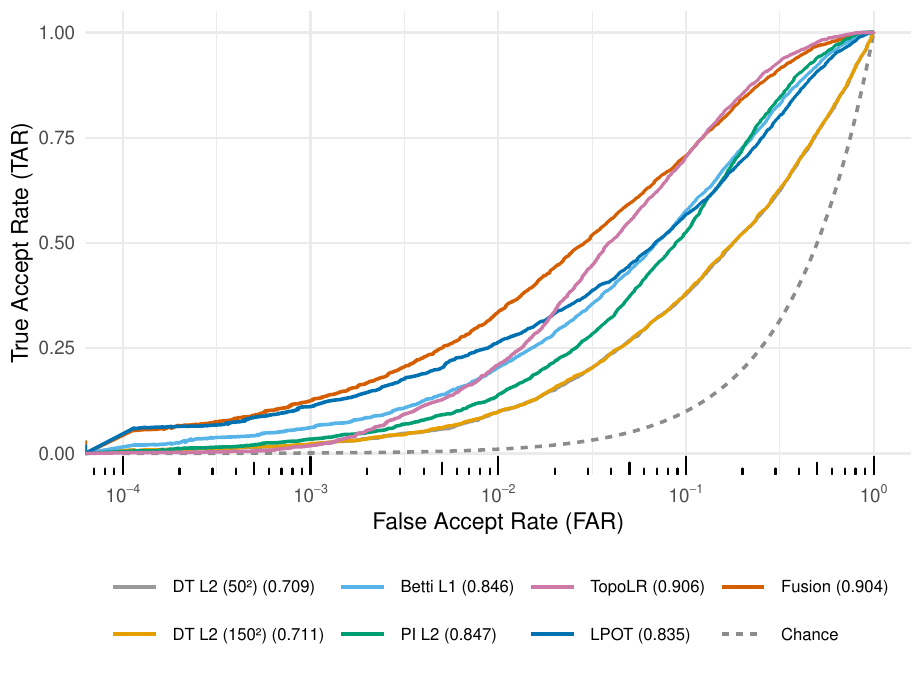}
\caption{ROC curves for all seven verification methods, computed by pooling test-set scores across the five cross-validation folds (FAR axis on a logarithmic scale). The dashed grey line marks chance performance, where TAR $=$ FAR. The legend reports AUC values computed on the pooled scores, which are shown only for this visualization; Table~\ref{tab:main} reports the primary fold-mean AUC estimates and may differ by small amounts because of the per-fold averaging convention. TopoLR achieves the highest pooled AUC, but the curves for LPOT and Fusion cross it in the low-FAR region, reflecting LPOT's advantage at stringent operating points.}
\label{fig:roc}
\end{figure}

Across all seven methods, topological summaries decisively outperform pixel-level geometry, and TopoLR and LPOT, the component methods used in fusion, show complementary behavior at the strictest operating points. We establish these results on FVC2000 DB1~\cite{maio2002fvc} under identity-disjoint 5-fold cross-validation: we partition the 100 identities into 5 groups of 20, and each fold trains on 4 groups and tests on the held-out group, so no identity appears in both training and test within any fold. An image-level split would allow a model to learn specific fingers during training and be evaluated on other impressions of those same fingers, a form of leakage that would inflate performance estimates. Within each fold, all possible image pairs yield 204{,}480 training pairs and 12{,}720 test pairs, each labeled genuine or impostor according to whether its two images come from the same finger. We report the area under the ROC curve (AUC), the equal error rate (EER), and the true accept rate at fixed false accept rates (TAR@FAR), with all fold-specific thresholds, score normalizations, learned model coefficients, and fusion weights estimated from training data only. Throughout, an AUC near 1 indicates near-perfect discrimination, an AUC of 0.5 indicates chance performance, and a lower EER is better; we report TAR at the stringent FAR values most relevant to security. Full protocol details, including a statistical caveat on fold-to-fold variance at the lowest FAR, appear in \emph{Materials and Methods} and \emph{SI Appendix}, section~\ref{si:eval}. Table~\ref{tab:main} presents results for all seven methods, and Fig.~\ref{fig:roc} shows pooled ROC curves.

\begin{table}[!ht]
\centering
\caption{Main comparison of all seven verification methods (mean $\pm$ SD across 5 folds, where SD is the standard deviation). EER and TAR values are percentages.}
\label{tab:main}
\resizebox{\linewidth}{!}{%
\begin{tabular}{lcccccc}
\toprule
Method & AUC & EER & TAR@$10^{-3}$ & TAR@$2{\times}10^{-3}$ & TAR@$5{\times}10^{-3}$ & TAR@$10^{-2}$ \\
\midrule
DT Down.\ $L^2$ ($50^2$) & .710$\pm$.024 & 34.5$\pm$2.6 & 1.7$\pm$0.8 & 3.0$\pm$1.4 & 5.9$\pm$2.5 & 9.5$\pm$4.0 \\
DT Down.\ $L^2$ ($150^2$) & .712$\pm$.024 & 34.4$\pm$2.7 & 1.9$\pm$0.6 & 3.1$\pm$1.2 & 6.0$\pm$2.6 & 9.6$\pm$3.7 \\
Betti $L^1$ & .847$\pm$.015 & 23.7$\pm$1.7 & 6.6$\pm$0.9 & 8.8$\pm$0.8 & 13.9$\pm$1.6 & 19.9$\pm$2.3 \\
PI $L^2$ & .847$\pm$.015 & 23.0$\pm$1.7 & 3.0$\pm$0.5 & 4.6$\pm$0.4 & 8.7$\pm$1.0 & 13.6$\pm$1.8 \\
TopoLR & .906$\pm$.011 & 17.4$\pm$1.6 & 4.9$\pm$4.2 & 8.3$\pm$5.6 & 14.9$\pm$6.7 & 23.0$\pm$7.4 \\
LPOT & .836$\pm$.008 & 25.2$\pm$1.2 & 11.2$\pm$4.8 & 13.9$\pm$5.8 & 19.8$\pm$5.7 & 25.6$\pm$6.7 \\
Fusion & .906$\pm$.008 & 17.8$\pm$1.3 & 13.4$\pm$5.7 & 18.5$\pm$6.1 & 27.0$\pm$7.5 & 35.6$\pm$7.1 \\
\bottomrule
\end{tabular}}
\end{table}

\subsection{Topology outperforms geometry at matched resolution}

Both geometry-only baselines perform substantially worse than all topological methods. The reduced-resolution baseline achieves AUC 0.710, while the matched-resolution baseline achieves AUC 0.712. Even when comparing at the same $150 \times 150$ resolution used by the topological methods, the geometry baseline falls well short of the global topological summaries (AUC 0.847), confirming that the performance advantage of topology is not an artifact of resolution differences. The distance transform itself is the same input in both cases; the difference lies in how that input is summarized, pixel-by-pixel comparison versus topological analysis. Both baselines compare unaligned images, an invariance confound we examine under \emph{Limitations}.

Having established that topology beats geometry, we next compare the two topological summaries with each other. Both global topological baselines achieve AUC 0.847 and similar EER values of 23.7\% and 23.0\% respectively. Despite these aggregate similarities, they differ substantially at low FAR: Betti curves achieve TAR@$10^{-3}$ of 6.6\% versus 3.0\% for persistence images. Because the two methods differ simultaneously in dimensionality, distance metric, and kernel parameters, we projected each representation onto its own 50-dimensional basis estimated by principal component analysis on training-fold features and scored test pairs with the same $L^2$ metric. The AUC values change little relative to fold-to-fold variability and the low-FAR gap persists (\emph{SI Appendix}, section~\ref{si:pca} and Table~\ref{tab:pca}), suggesting that the difference reflects the representations themselves: at strict thresholds, the count-based Betti summary appears more resistant to noise in the extreme tail of the impostor distribution than the smoothed joint birth--persistence distribution.

\subsection{Learned and matching-based methods}

Among the seven primary methods in Table~\ref{tab:main}, the full TopoLR model is the strongest non-fusion method by AUC (0.906) and EER (17.4\%), showing that learning to combine global and local topological features substantially improves overall discrimination. Ablation studies (\emph{SI Appendix}, section~\ref{si:ablation} and Table~\ref{tab:ablations}) show that most of this improvement comes from the learned combination weights rather than the local features: global-only TopoLR already achieves AUC 0.904, while adding local Betti information contributes a further 0.002 that is well within the fold-to-fold noise.

LPOT presents an intriguing contrast. Its AUC of 0.836 and EER of 25.2\% are comparable to, or slightly below, the global baselines. Yet at low FAR, LPOT achieves the highest TAR among individual methods: 11.2\% at FAR $= 10^{-3}$, compared to 6.6\% for the Betti baseline and 4.9\% for TopoLR. Fold-to-fold variability is substantial at this operating point, with standard deviations of 4--5 percentage points, so we urge caution in interpreting the magnitudes. In the fold means, LPOT exceeds TopoLR at all reported TAR thresholds; at FAR $= 10^{-3}$, LPOT also exceeds TopoLR in all five folds. This suggests that spatially coherent local matching is particularly effective at separating the hardest impostor pairs, those that are globally similar but differ in local spatial arrangement. The ablations support this reading: removing LPOT's geometry cost collapses TAR@$10^{-3}$ from 11.2\% to 2.3\%, and removing the dustbin reduces it from 11.2\% to 7.5\% (\emph{SI Appendix}, Table~\ref{tab:ablations}).

Fusion achieves the best mean TAR at every reported low-FAR operating point. At TAR@$10^{-2}$, where relative fold-to-fold variability is lower, Fusion achieves 35.6\% compared to 25.6\% for LPOT and 23.0\% for TopoLR. At the more stringent TAR@$10^{-3}$, Fusion achieves 13.4\% compared to 11.2\% for LPOT and 4.9\% for TopoLR, though fold-to-fold variability is large at this threshold; Fusion exceeds both component methods in every fold at both operating points. Fusion's AUC of 0.906 matches TopoLR's, and its EER of 17.8\% is comparable to TopoLR's 17.4\%, the small difference being well within fold-to-fold standard deviation. Fusion therefore inherits TopoLR's overall discrimination while gaining LPOT's substantial advantage at strict thresholds, consistent with the two methods capturing complementary aspects of fingerprint identity.

\section{Discussion}

Together, the results show that topology improves discrimination by summarizing ridge geometry while discarding nuisance variation, and that global and local topological scores fail in different parts of the impostor distribution.

\subsection{The discriminative power of topology}

The most fundamental finding of this work is that topological representations substantially outperform pure geometry for fingerprint verification. The AUC improvement from the geometry baselines to the global topological methods arises from summarizing the distance transform through persistent homology rather than using raw pixel values, and it holds at matched resolution. Why should topology be discriminative for fingerprints? The geometric argument developed above provides a partial answer: the $H_1$ persistence diagram encodes the birth and death of loops as the expanding tube around the ridge set encloses valley regions, and this process is governed by ridge spacing, branching, and continuity, precisely the features that vary between different fingers. But this argument only establishes that the persistence diagram \emph{could} differ between fingers; it does not guarantee that the differences are large enough to be useful for verification, or that they survive the noise and deformation inherent in repeated impressions of the same finger. Our results provide empirical evidence that both conditions hold: topological summaries differ sufficiently across identities while remaining sufficiently stable across impressions of the same finger. The stability theorem for persistence diagrams provides a theoretical basis for the latter observation, at least for the component of impression variability that manifests as small pointwise perturbations of the distance transform. The advantage is not that topology encodes \emph{more} information than the pixel grid, but that it discards nuisance variation, spatial rearrangement and local noise, while retaining discriminative structure: ridge spacing, branching, and enclosure patterns.

\subsection{The complementarity of TopoLR and LPOT}

\begin{figure}[!ht]
\centering
\includegraphics[width=\linewidth]{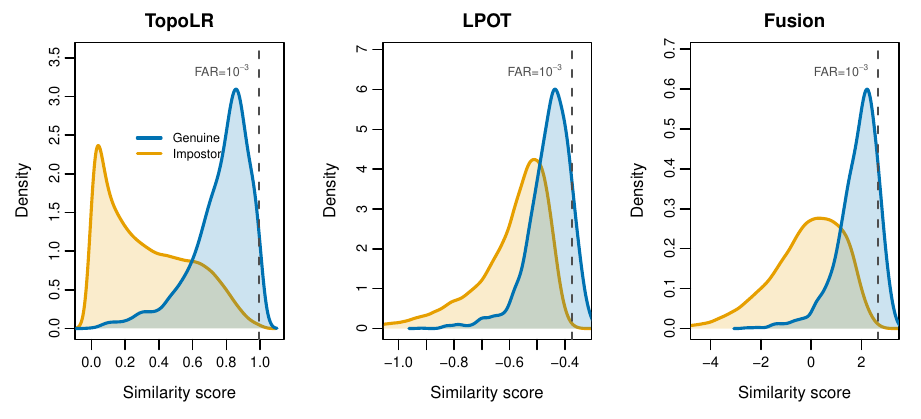}
\caption{Score distributions for TopoLR, LPOT, and their fusion, with impostor pairs (orange) and genuine pairs (blue). The dashed line is a pooled visualization threshold, set at the 99.9th percentile of pooled impostor scores; the reported TAR values in Table~\ref{tab:main} use fold-specific thresholds estimated from training impostor scores, so this dashed line is illustrative rather than the operational threshold. TopoLR achieves better overall separation but LPOT places more genuine-pair mass above the strict threshold, explaining its superior TAR at low FAR. Fusion inherits LPOT's low-FAR advantage and improves upon it while maintaining essentially the same mean AUC as TopoLR in the fold-level results reported in Table~\ref{tab:main}.}
\label{fig:scores}
\end{figure}

TopoLR and LPOT capture different aspects of fingerprint identity and excel in different operating regimes (Fig.~\ref{fig:scores}). TopoLR learns a global similarity function from Betti curves, capturing aggregate properties of the ridge pattern: how many loops exist at each scale, how these counts vary across the image, and how these quantities differ between genuine and impostor pairs. LPOT achieves lower AUC and higher EER, yet substantially higher TAR at low FAR. How can a method with worse average discrimination achieve better performance at the strictest operating points? The answer lies in the shape of the score distributions (Fig.~\ref{fig:scores}). At low FAR, the system sets the threshold to reject all but a tiny fraction of impostors, the hardest cases that receive the highest scores. By requiring that matched patches be not only topologically similar but also spatially consistent, LPOT rejects impostor pairs that happen to share similar global topological profiles but differ in the spatial arrangement of their local features.

The fusion of these methods achieves TAR@$10^{-3}$ of 13.4\%, exceeding both methods individually, with only a single scalar weight tuned on each fold's training set. Across the five folds the selected weights are $w = 0.42, 0.38, 0.40, 0.58, 0.64$ (mean $0.48$, SD $0.12$), with both methods receiving substantial weight in every fold. If the two methods were redundant, fusion would provide little benefit beyond the better of the two. The consistent gains across all five folds support the interpretation that TopoLR and LPOT capture complementary predictive structure: aggregate topological statistics on one hand, and spatially structured local correspondences on the other. The large fold-to-fold variance at TAR@$10^{-3}$ warrants caution in interpreting the magnitude of the fusion improvement at this operating point; the complementarity is better supported at TAR@$10^{-2}$, where fusion achieves 35.6\% versus 25.6\% (LPOT) and 23.0\% (TopoLR) with lower relative variance.

\subsection{Limitations}

Six limitations bound these conclusions. First, variance at low FAR is substantial. Standard deviations of roughly 4--6 percentage points at TAR@$10^{-3}$ reflect the statistical challenge of estimating performance when only about 12 impostor pairs per fold are expected to exceed the threshold; readers should place more confidence in TAR@$10^{-2}$, and applications requiring precise estimates at very low FAR would need substantially larger evaluation datasets (\emph{SI Appendix}, section~\ref{si:eval}).

Second, we have not optimized hyperparameters. We fixed all parameter values through preliminary exploration on the same dataset rather than a held-out development set, so we cannot rule out dataset-level selection bias even though fold-level training/test separation is strict. Different parameter choices could change the relative rankings of methods. The appropriate next step is not exhaustive tuning but targeted stress tests: verifying that modest changes to key parameters do not qualitatively alter the conclusions.

Third, all results are from a single dataset acquired with a single low-cost optical sensor. The FVC competitions provide additional benchmarks with different sensors and image qualities~\cite{maio2002fvc}; testing on these datasets, using the same parameter values without re-tuning, would assess whether our methods generalize across acquisition conditions.

Fourth, we have not stratified results by fingerprint pattern type (arch, loop, whorl). The relationship between pattern type and $H_1$ features is not straightforward: curving ridges do not by themselves create enclosed valley regions, and the $H_1$ signal is tied to genuine enclosures and to the image boundary's role in determining which valley regions are enclosed. If pattern types differ in the number and arrangement of such enclosures, global topological summaries may be more discriminative for separating different pattern types than for distinguishing two fingers with the same pattern type. Whether this meaningfully affects verification performance remains an open empirical question.

Fifth, the geometry-only baselines compare raw distance-transform images without any alignment, while global topological summaries are by construction invariant to rigid translation and, in the continuous limit, to rotation. They are \emph{not} invariant to elastic deformation, which changes ridge spacing and hence the distance transform. LPOT, by contrast, uses centered spatial coordinates in its geometry cost and is therefore sensitive to rotation and non-rigid deformation, though its centering step provides translation invariance. Our evaluation therefore conflates two effects: the intrinsic value of topological versus geometric information, and the invariance and stability advantages that topological summaries enjoy over unaligned pixelwise comparison. An informative control would be an aligned geometry baseline, for example maximizing normalized cross-correlation over a grid of translations and rotations before computing $L^2$ distance; we leave this comparison to future work (\emph{SI Appendix}, section~\ref{si:background} discusses the relevant invariance properties in detail).

Finally, the best EER among the seven primary methods (17.4\%) remains well above the error rates achieved by mature minutiae-based systems on comparable benchmarks~\cite{maio2002fvc,maltoni2022handbook}. This gap reflects not a limitation of topology per se but rather a consequence of our deliberately constrained setting: we excluded minutiae extraction, skeletonization, and explicit alignment, the very operations that drive conventional performance.

\subsection{Outlook}

The most immediate direction for future work is to evaluate whether topological features complement minutiae-based scores. Our methods deliberately exclude the features that drive conventional systems; if the information they capture is partly independent of minutiae-based scores, fusing topological and minutiae-based scores could improve performance beyond either alone. A second direction is richer topological representations: $H_0$ features could be informative under alternative filtrations, scalar fields other than the distance transform, such as ridge orientation coherence or local ridge frequency, could yield complementary signatures, and multi-parameter persistence could capture relationships between fields. Third, applying the same pipeline with fixed parameters to additional datasets would test generalizability, a prerequisite for practical deployment. Finally, the connection between topological representations and fingerprint pattern type merits investigation: if arches, loops, and whorls produce characteristically different persistence diagrams, understanding this relationship could lead to topologically informed pattern classification or to verification methods that adapt their comparison strategy to the inferred pattern type. The entire pipeline presented here, from distance transform through topological summary to verification score, is mathematically principled, fully specified by a modest set of fixed parameters, and exactly reproducible, providing a foundation for these investigations.

\section{Materials and Methods}
\subsection{Dataset and preprocessing}
FVC2000 DB1 contains 800 images from 100 identities with 8 impressions of the same finger each, acquired at 500 dpi with a low-cost optical sensor at $300 \times 300$ pixels~\cite{maio2002fvc}. The impressions vary in pressure, moisture, rotation, and partial capture. We preprocessed raw images using the enhancement algorithm of Hong, Wan, and Jain~\cite{hong1998fingerprint}, as implemented in the Fingerprint-Enhancement-Python library~\cite{deshmukh2018fingerprint} with default parameters, to produce binary ridge/valley segmentations; the enhancement step is not part of our contribution, and we treat the binary images as given. On loading each image, we verify that ridge pixels constitute fewer than half of all pixels, as expected from the standard convention, and invert the labeling if not.

\subsection{Distance transforms and downsampling}
We compute the two distance transform fields (Eqs.~\ref{eq:dt_ridge} and \ref{eq:dt_valley}) on the full-resolution binary image and then downsample by selecting every $k$-th pixel, which keeps persistent homology computation tractable across the full dataset: to $150 \times 150$ for all persistent homology computations and one baseline, and to $50 \times 50$ for the coarser baseline. Applying the distance transform before downsampling ensures that values reflect full-resolution ridge geometry, mitigating aliasing and connectivity artifacts. We verified on a random sample of 10 images that the $H_1$ persistence diagrams of both the ridge and valley distance transforms at $300 \times 300$ and $150 \times 150$ are close in bottleneck distance (\emph{SI Appendix}, section~\ref{si:rep}).

\subsection{Persistent homology computation}
We compute sublevel set persistent homology on the cubical complex defined by the pixel grid using the GUDHI library~\cite{maria2014gudhi} via the R package TDA~\cite{fasy2014tda}. Function values are provided on pixels and extended to lower-dimensional cells by minimum over incident pixels, producing a valid sublevel filtration; all reported birth and death scales correspond to pixel values. Before computation we clamp distance transform values to $[0, 20]$ and round to one decimal place; the clamping threshold well exceeds the maximum inter-ridge distance in the dataset, and the rounding discretizes the filtration to 201 levels with at most 0.05 pixels of error (\emph{SI Appendix}, section~\ref{si:rep}).

\subsection{Vectorizations}
Betti curves are evaluated at 50 equally spaced filtration values on $[0,20]$ per channel and normalized by ridge area fraction, with a minimum of 0.01 to avoid division by small numbers. Global persistence images use a $20 \times 20$ grid with Gaussian bandwidth $\sigma = 0.5$ (800 dimensions per image after concatenating channels); local persistence images use a $30 \times 30$ grid with $\sigma = 0.25$ (1{,}800 dimensions per anchor). We $L^2$-normalize the global persistence-image vector after concatenating its ridge and valley channels; for local LPOT descriptors we $L^2$-normalize each channel separately before concatenation, with no further normalization. Local patches use 64 anchor points sampled uniformly at random from ridge pixels with a deterministic image-specific seed, a circular patch of radius 12 pixels cropped from the full-image distance transforms, and a disk mask assigning exterior pixels the maximum filtration value (\emph{SI Appendix}, section~\ref{si:rep}).

\subsection{TopoLR training}
We train $L^2$-regularized logistic regression (regularization strength $\lambda = 0.01$) by mini-batch stochastic gradient descent with balanced sampling with replacement: each batch contains 5{,}000 genuine and 5{,}000 impostor pairs, with learning rate 0.1 decaying by a factor of 0.9 every 20 iterations for 200 iterations. Pair features are standardized using means and standard deviations estimated from a random sample of 20{,}000 training pairs, applied unchanged to test pairs.

\subsection{LPOT solver}
The augmented transport problem fixes the dustbin cost at 0.4, the dustbin mass budget at 20\% of the total, and the geometry weight at $\beta = 6$, all set during preliminary exploration and held fixed across folds. We solve with the Sinkhorn algorithm using entropic regularization parameter 0.12 and exactly 60 iterations with no early stopping, ensuring exact reproducibility.

\subsection{Evaluation protocol}
We use identity-disjoint 5-fold cross-validation with a fixed random seed. For TAR@FAR, we set thresholds using only the training impostor score distribution: to target a false accept rate of $\alpha$, we find the threshold above which only a fraction $\alpha$ of training impostor scores fall, then apply this threshold to test genuine scores. The fusion weight $w$ is selected by grid search over $[0,1]$ in increments of 0.02 on training data. We compute all metrics per fold and report mean $\pm$ standard deviation across folds; ROC curves pool test scores across folds for visualization only. Nominal operating-point labels such as FAR $= 10^{-3}$ are approximate targets rather than exact realized rates, and fold-to-fold variance at the lowest FAR is correspondingly high (\emph{SI Appendix}, section~\ref{si:eval}).

\subsection{Software and reproducibility}
We implement the pipeline in R using EBImage~\cite{pau2010ebimage} for image input/output and distance transforms; TDA~\cite{fasy2014tda} with GUDHI~\cite{maria2014gudhi} for persistent homology; TDAvec~\cite{islambekov2025tdavec} for persistence images; pROC~\cite{robin2011proc} for ROC analysis; and future and future.apply~\cite{bengtsson2021future} for parallel computation. We fixed all random seeds and iteration counts so that the entire pipeline produces identical results across runs. Runtime measurements appear in \emph{SI Appendix}, section~\ref{si:runtime}.

\section*{Acknowledgments}
This material is based upon work supported by the National Science Foundation under Grant No.~DMS-2424556 while the authors were in residence at the Institute for Computational and Experimental Research in Mathematics in Providence, RI, during the Collaborate@ICERM program ``Fingerprint Matching via Topological Data Analysis.'' LZ was supported in part by grant NSF BCS-2318171. NAM was supported in part by San Francisco State University institutional start-up
funds.

\section*{Author Contributions}
Conceptualization, C.M.T., N.A.M., E.M., Z.S., and L.Z.; Data curation, C.M.T.; Formal analysis, C.M.T.; Funding acquisition, C.M.T., N.A.M., E.M., Z.S., and L.Z.; Investigation, C.M.T., N.A.M., E.M., and L.Z.; Methodology, C.M.T., N.A.M., E.M., and L.Z.; Software, C.M.T.; Writing -- original draft, C.M.T.; and Writing -- review \& editing, C.M.T., N.A.M., E.M., Z.S., and L.Z.

\section*{Competing Interests}
The authors declare no competing interest.

\section*{Data Availability}
All analysis code, together with the scripts that reproduce every figure and table in this article and instructions for obtaining and preparing the benchmark data, is available in a public repository at \url{https://github.com/chadtopaz/fingerprint-tda} and archived at Zenodo (\url{https://doi.org/10.5281/zenodo.20772749}). The FVC2000 DB1 benchmark dataset is available from the Fingerprint Verification Competition~\cite{maio2002fvc}.

\clearpage
\setcounter{section}{0}\setcounter{subsection}{0}\setcounter{figure}{0}\setcounter{table}{0}\setcounter{equation}{0}
\renewcommand{\thesection}{S\arabic{section}}
\renewcommand{\thesubsection}{\thesection.\arabic{subsection}}
\renewcommand{\thefigure}{S\arabic{figure}}
\renewcommand{\thetable}{S\arabic{table}}
\renewcommand{\theequation}{S\arabic{equation}}
\section*{Supporting Information}
This appendix provides extended background, full methodological detail, and supporting analyses for the main text. Section~\ref{si:background} reviews friction ridge biology, conventional fingerprint recognition, and topological data analysis. Section~\ref{si:eval} details the evaluation framework and the statistical behavior of low-FAR estimates. Section~\ref{si:rep} gives the full topological representation pipeline. Section~\ref{si:methods} specifies the verification methods and collects key parameter values (Table~\ref{tab:params}). Section~\ref{si:ablation} reports ablation studies (Table~\ref{tab:ablations}), Section~\ref{si:runtime} reports runtime (Table~\ref{tab:runtime}), and Section~\ref{si:pca} reports a dimension-matched control comparison (Table~\ref{tab:pca}). Equation, figure, and table numbers in this appendix are prefixed by S and are independent of the main text.

\section{Extended background}
\label{si:background}

This section expands the brief treatment in the main text. The fingerprint material follows refs.~\cite{cummins1961finger, babler1991embryologic, AFIS2011, daluz2018fundamentals, maltoni2022handbook} and the reviews \cite{yager2004fingerprint, bansal2011minutiae, PERALTA201567, ali2016overview, valdes2019review, win2020fingerprint, priesnitz2021overview}; the TDA material follows refs.~\cite{otter2017roadmap, munch2017user, wasserman2018topological, carlsson2020topological, amezquita2020shape, skaf2022topological}.

\subsection{Friction ridge patterns and fingerprints}

\emph{Friction ridges} are raised patterns on the skin of human fingers, palms, and soles. Their development begins with cell differentiation in the fetal hand around six weeks after fertilization. Swellings called \emph{volar pads} form on the fingertips, and primary friction ridges begin forming when these pads recede around weeks 10--11 of gestation. The epidermal layers thicken by weeks 14--15, and secondary ridges appear between weeks 15 and 17. The process is largely complete by 25 weeks of gestation, with the newly formed ridges persisting throughout an individual's life barring severe injury. The particular arrangement of ridges into patterns depends on genetic makeup, maternal nutrition, the intrauterine environment, and movements of the fetus.

There are three primary pattern types. \emph{Loops}, the most common, are characterized by ridges that enter from one side, curve, and exit on the same side. \emph{Whorls} form circular or spiral patterns. \emph{Arches}, the simplest type, consist of ridges running from one side to the other without backtracking. Each type has subtypes: loops can be ulnar or radial, arches can be plain or tented, and so on. Patterns are further characterized by ridge count, the number of ridges between two reference points. Other important features include \emph{deltas}, triangular regions near the divergence of ridge flow lines, and cores, the innermost turning points of loops. The finest details are called \emph{minutiae} and include ridge endings, bifurcations, short ridges, dots, and lakes. A \emph{ridge ending} is where a ridge abruptly terminates. A \emph{bifurcation} is where a single ridge splits into two. A \emph{short ridge} begins, travels a short distance, and ends. A \emph{dot} is an isolated ridge fragment so small that it appears as a roughly circular point. A \emph{lake} is a single ridge that bifurcates and then rejoins. While conventional wisdom holds that every ridge pattern is unique, the capacity of humans and computers to assess the degree of difference is finite, and one recent study found that fingerprints from different fingers of the same person share strong similarities~\cite{guo2024unveiling}.

Ridge patterns and fingerprints differ in an important way. Ridge patterns are inherent features of the skin, while \emph{fingerprints} are variable reproductions whose quality depends on capture method, pressure, medium, and surface condition. \emph{Known} or \emph{exemplar} fingerprints are captured directly from subjects, traditionally by rolling each inked finger across a tenprint card and now by Livescan devices that scan fingerprints optically or capacitively, eliminating the messiness and variability of ink-based methods while enabling rapid exchange across agencies. \emph{Latent} prints are unintentionally deposited and often invisible, requiring recovery techniques chosen based on surface characteristics: porous surfaces such as paper stabilize prints by absorbing substances, while non-porous surfaces such as glass leave prints more vulnerable to smudging and environmental degradation. The recovery and analysis of latent prints require careful consideration of substrate, composition of the deposit, and the potential for distortion from skin elasticity and pressure variations.

\emph{Fingerprint verification} (1:1 matching) authenticates an individual by comparing a presented fingerprint against a stored template, while \emph{fingerprint identification} (1:N matching) searches a database to determine who left a given print. Automated Fingerprint Identification Systems (AFIS) support both tasks; their computational infrastructure has evolved considerably since its introduction in the 1970s, but the core algorithmic paradigm has remained largely unchanged. Specific implementations include the FBI's Next Generation Identification system, successor to the earlier Integrated AFIS and one of the largest biometric databases in the world, as well as systems from vendors such as NEC, Thales, and Idemia. In law enforcement, AFIS supports tenprint-to-tenprint searches that check for a prior record, latent-to-tenprint searches that identify who left a crime scene print, latent-to-latent searches that link multiple crime scenes, and tenprint-to-latent searches that match new arrestee prints against unresolved evidence.

\subsection{Computational approaches and their limitations}

The computational fingerprint analysis literature is vast; the reviews~\cite{yager2004fingerprint, bansal2011minutiae, PERALTA201567, ali2016overview, valdes2019review, win2020fingerprint, priesnitz2021overview} collectively reference hundreds of individual studies. We give a high-level overview, emphasizing the aspects most relevant to our approach. Standard automated pipelines proceed through several stages. \emph{Enhancement} improves image quality using contextual filters, often Gabor filters tuned to local ridge orientation and frequency, along with local contrast enhancement. \emph{Binarization} converts the enhanced grayscale image to a binary representation of ridges and valleys. \emph{Skeletonization} thins the binary ridges to single-pixel width. \emph{Minutiae detection} scans the thinned image to identify ridge endings and bifurcations based on pixel connectivity: a ridge pixel with exactly one neighbor signals an ending, while one with three neighbors signals a bifurcation. Beyond minutiae, some algorithms extract additional features such as sweat pores, ridge counts between landmarks, and local quality measures. \emph{Matching} aligns two representations by their features and computes a similarity score based on the number and spatial arrangement of corresponding minutiae, and \emph{indexing} organizes the database for efficient retrieval. The specific algorithms employed by operational AFIS vary across systems and vendors, and the details often remain proprietary.

Current systems face well-documented limitations. Accuracy depends critically on image quality, and algorithms must accommodate variations in capture angle, pressure, moisture, and elastic skin deformation. More fundamentally, any score-based system, including the topological methods we propose, requires a threshold to convert a continuous score into a binary accept/reject decision, and this threshold reflects application-specific security requirements rather than an intrinsic property of the comparison.

\subsection{Topological data analysis}
\label{si:tda}

\emph{Topology} studies qualitative properties of spaces that are preserved under continuous deformations such as stretching and bending, but not tearing or gluing. \emph{Topological data analysis} applies these ideas to extract structural information from data by studying its shape at multiple scales, and has been applied across areas including cancer biology~\cite{Nicolau2011}, biological aggregations~\cite{topaz2015topological}, granular materials~\cite{bassett2012influence}, voting~\cite{feng2021persistent}, and global development~\cite{Banman2018}.

A central tool is \emph{persistent homology}, which tracks topological features across a range of scales. The input is a \emph{filtration}, a nested sequence of topological spaces parameterized by a scale parameter. Different data types give rise to different filtration constructions. For point cloud data, the Vietoris--Rips construction connects pairs of points within distance $\epsilon$ and fills in higher-dimensional simplices whenever all pairwise distances are at most $\epsilon$. For scalar fields defined on a grid, the setting relevant to our work, the \emph{sublevel set filtration} provides a natural alternative: given a function $f$, the sublevel set at scale $t$ is $L_t(f) = \{x : f(x) \leq t\}$, which grows monotonically as $t$ increases. For a function on a pixel grid, this filtration is realized on a \emph{cubical complex} in which the cells are pixels (2-cubes), edges between adjacent pixels (1-cubes), and vertices at pixel corners (0-cubes).

\emph{Homology} quantifies the topology of each space via Betti numbers. In the two-dimensional setting, $\beta_0$ counts connected components, $\beta_1$ counts one-dimensional loops that do not bound a filled region, and Betti numbers in dimensions two and higher are zero. We write $H_0$ and $H_1$ for the corresponding homology groups. \emph{Persistent homology} tracks how these features evolve as the filtration parameter increases. A connected component is born when a new region of the sublevel set appears and dies when it merges with an older component; a loop is born when a cycle forms and dies when the region it encloses is completely filled in. The collection of all birth--death pairs is recorded in a \emph{persistence diagram}, a multiset of points $(b, d)$ above the diagonal $d = b$. The \emph{persistence} $d - b$ is the range of scales over which a feature persists; features near the diagonal are often attributable to noise.

A key property of persistence diagrams is \emph{stability}: if two functions $f$ and $g$ on the same domain satisfy $\|f - g\|_\infty \leq \delta$, then the bottleneck distance between their persistence diagrams is at most $\delta$~\cite{otter2017roadmap}. For fingerprint verification, this means that minor imaging variations, slight differences in ink density, small amounts of noise, or local smudging, that produce small pointwise changes in the distance transform will produce correspondingly small changes in the persistence diagram. A related property is rotation invariance. For a continuous function on $\mathbb{R}^2$, the Euclidean distance transform and hence the persistence diagram are exactly rotation-invariant. On a discrete pixel grid, however, both the distance transform and the cubical complex break continuous rotational symmetry, so persistence diagrams of rotated images are only approximately equal, with discrepancies that depend on grid resolution and rotation angle. In our setting, a $150 \times 150$ grid with moderate rotations between impressions, this discretization effect is small relative to other sources of variation but is not zero. The stability guarantee does not extend to large geometric deformations such as elastic stretching of the skin, which can produce large pointwise changes in the distance transform; robustness to such deformations is an empirical property of our methods, not a theoretical guarantee.

\section{Evaluation framework}
\label{si:eval}

Given a pair of fingerprint images, a verification system produces a similarity score; higher scores indicate greater confidence that the two images come from the same finger. A \emph{genuine} pair consists of two images from the same finger, while an \emph{impostor} pair consists of images from different fingers. By varying a threshold on the score, one trades off between the \emph{false accept rate} (FAR), the fraction of impostor pairs incorrectly accepted, and the \emph{false reject rate} (FRR), the fraction of genuine pairs incorrectly rejected.

We use 5-fold cross-validation with folds constructed at the \emph{identity} level: we partition the 100 identities into 5 groups using a fixed random seed, with no identity appearing in both training and test within any fold. Splitting at the image level would let a model learn to recognize specific fingers during training and then be evaluated on other impressions of those same fingers, a form of leakage that would inflate performance estimates. Within each fold, we generate all possible image pairs: 204{,}480 training pairs from 80 identities and 12{,}720 test pairs from 20 identities, each labeled genuine or impostor.

We report three families of metrics. The \emph{receiver operating characteristic} (ROC) curve plots the true accept rate against the false accept rate as the decision threshold varies; the \emph{area under the ROC curve} (AUC) summarizes overall discrimination, with 1.0 indicating perfect separation and 0.5 chance. The \emph{equal error rate} (EER) is the error rate at the threshold where FAR equals FRR. The \emph{true accept rate at a fixed false accept rate} (TAR@FAR) measures the fraction of genuine pairs accepted when the threshold is calibrated to a target FAR on training data; we report TAR at nominal FAR values of $10^{-3}$, $2 \times 10^{-3}$, $5 \times 10^{-3}$, and $10^{-2}$. We compute all metrics per fold and report mean $\pm$ standard deviation across folds.

We enforce strict separation between training and test data within each fold. For learned methods, we estimate all learned model parameters from training data only and apply them unchanged to test data. For TAR@FAR computation, we set thresholds using only the training impostor score distribution: to target a false accept rate of $\alpha$, we find the threshold above which only a fraction $\alpha$ of training impostor scores fall, then apply this threshold to test genuine scores to measure the true accept rate. For the fusion method, we tune the combination weight on training data only. ROC curves pool test scores across all five folds for visualization, but the reported AUC values are per-fold averages, not computed on pooled scores.

\subsection{A statistical caveat at low FAR}

Although TAR@FAR thresholds are set from training impostor scores rather than test scores, the \emph{measured TAR} at these thresholds varies across folds for two reasons: the training-derived threshold itself varies because each fold's training impostor distribution differs, and the test genuine scores used to compute TAR are drawn from different identities across folds. A separate issue is that the operating-point label, for example ``$\text{FAR} = 10^{-3}$,'' is only nominal: with approximately 12{,}000 test impostors per fold, the realized test FAR can deviate substantially from $10^{-3}$, so the label should be understood as an approximate target rather than an exact realized rate. Together these effects make TAR@$10^{-3}$ less stable in relative terms: for the strongest methods, fold-to-fold standard deviations are roughly $\pm$4--6 percentage points on means of only 5--13 percentage points. We report these values because low-FAR performance is operationally important, but readers should interpret them with appropriate caution and give more weight to TAR@$10^{-2}$ for stable comparisons.

\section{Topological representation: full detail}
\label{si:rep}

\subsection{Distance transforms and downsampling}

The distance transform is a natural choice for several reasons. First, it converts a discrete binary pattern into a real-valued scalar field, the required input for sublevel set persistent homology. Second, it encodes geometric information, ridge width and spacing, in a natural way: the maximum value of $\mathrm{DT}_{\mathrm{ridge}}$ in a valley region reflects the half-width of that valley, and the medial axis of the valley reflects the shape of the bounding ridges. Third, the distance transform is simple and parameter-free: a binary image uniquely determines it.

We compute the distance transform fields on the full-resolution binary image and then downsample by selecting every $k$-th pixel in each dimension. FVC2000 DB1 images are $300 \times 300$ pixels. For one baseline method we downsample to $50 \times 50$; for all other methods, including all persistent homology computations, we use a $150 \times 150$ grid produced by selecting every second pixel. Downsampling makes cubical-complex persistent homology substantially faster while preserving the spatial scale of fingerprint ridge structure, whose characteristic widths span several pixels. Applying the distance transform before downsampling ensures that the values reflect full-resolution ridge geometry, mitigating the aliasing and connectivity artifacts that would arise from downsampling the binary image and recomputing the distance transform on the coarser grid. We verified on a random sample of 10 images that the $H_1$ persistence diagrams of both the ridge and valley distance transforms at $300 \times 300$ and $150 \times 150$ are close in bottleneck distance, with mean bottleneck 0.83 for the ridge channel and 0.68 for the valley channel, and maximum 1.00 for each, on the $[0, 20]$ filtration scale; the features eliminated under downsampling are predominantly low-persistence features of the finer grid.

\subsection{Sublevel set persistent homology}

We analyze each distance transform through sublevel set persistent homology on the cubical complex defined by the pixel grid, using the GUDHI library's cubical complex implementation via the R package TDA. We provide function values on the pixels (2-cubes), and GUDHI extends these values to lower-dimensional cells by assigning each edge or vertex the minimum value among its incident pixel cofaces, ensuring that every face appears no later than its cofaces and producing a valid sublevel set filtration. All birth and death scales correspond to pixel values. Before computation, we clamp distance transform values to the range $[0, 20]$ and round to one decimal place. The clamping threshold of 20 well exceeds the maximum inter-ridge distance in the dataset, where typical valley half-widths are 3--8 pixels at this resolution, ensuring that no meaningful ridge-valley structure is lost while preventing very large distances in background regions from dominating the filtration. The rounding discretizes the filtration to 201 distinct levels, reducing the number of topological events without materially affecting the persistence diagrams, since rounding to the nearest 0.1 introduces at most 0.05 pixels of error.

We do not use $H_0$ features in our scoring methods: in each distance-transform filtration, all connected components of the zero set, ridges for $\mathrm{DT}_{\mathrm{ridge}}$ and valleys for $\mathrm{DT}_{\mathrm{valley}}$, are born at $t = 0$ and merge rapidly as the sublevel-set tubes coalesce, producing a Betti curve that collapses to~1 within the first few filtration steps and offers little discriminative variation across identities. For the ridge-channel filtration, an $H_1$ feature's death scale is governed by the inradius of the enclosed valley region rather than by its area: a long, narrow valley can have a large area but a small inradius and hence a low death scale. The analogous statement holds for enclosed ridge regions tracked by the valley-channel filtration. The persistence $d-b$ depends on both this inradius and the birth scale, which is itself determined by the widest gap that must close before the loop forms. We compute persistent homology separately on the ridge and valley distance transforms, yielding two $H_1$ persistence diagrams per image. We use both because their sublevel filtrations grow from different sets: $\mathrm{DT}_{\mathrm{ridge}}$ grows outward from the ridge set and its $H_1$ features track valley regions enclosed by ridges, while $\mathrm{DT}_{\mathrm{valley}}$ grows outward from the valley set and its $H_1$ features track ridge regions enclosed by valleys. These two filtrations are not related by a simple reparameterization; they capture complementary aspects of the ridge--valley geometry.

\subsection{Betti curves and persistence images}

We evaluate the Betti curve on a grid of 50 equally spaced values from 0 to 20, yielding a 50-dimensional vector per diagram. Concatenating ridge and valley curves gives a 100-dimensional \emph{global Betti vector}. We normalize by ridge area fraction, the proportion of pixels that are ridges, with a minimum of 0.01 to avoid division by small numbers. Because images with greater ridge coverage naturally produce more topological features, this normalization ensures that the Betti vector reflects the shape of the topological profile rather than the overall amount of ridge content.

\emph{Persistence images}~\cite{adams2017persistence} proceed in three steps. First, we transform each point $(b,d)$ to \emph{birth--persistence coordinates} $(b, d - b)$. Second, we weight each point by a persistence-dependent function that vanishes at zero persistence, which downweights low-persistence features near the diagonal and is required for the stability of persistence images; in our TDAvec implementation we use the default linear ramp weighting. We then place a Gaussian kernel with bandwidth $\sigma$ at each weighted point and sum the kernels to create a smooth density over the birth--persistence plane. Third, we discretize this density on a regular grid and flatten it into a vector. We use two parameterizations: for the global persistence image baseline, a $20 \times 20$ grid with $\sigma = 0.5$, yielding 400 dimensions per diagram or 800 after concatenating ridge and valley channels; for local persistence images used in LPOT, a finer $30 \times 30$ grid with $\sigma = 0.25$, yielding 900 dimensions per channel per patch and 1{,}800 per anchor. For the global persistence-image baseline, we concatenate ridge and valley channels and $L^2$-normalize the resulting 800-dimensional vector; for local LPOT descriptors, we $L^2$-normalize each channel independently before concatenation, as detailed below.

\subsection{Local patch extraction}

\begin{figure}[!ht]
\centering
\includegraphics[width=\textwidth]{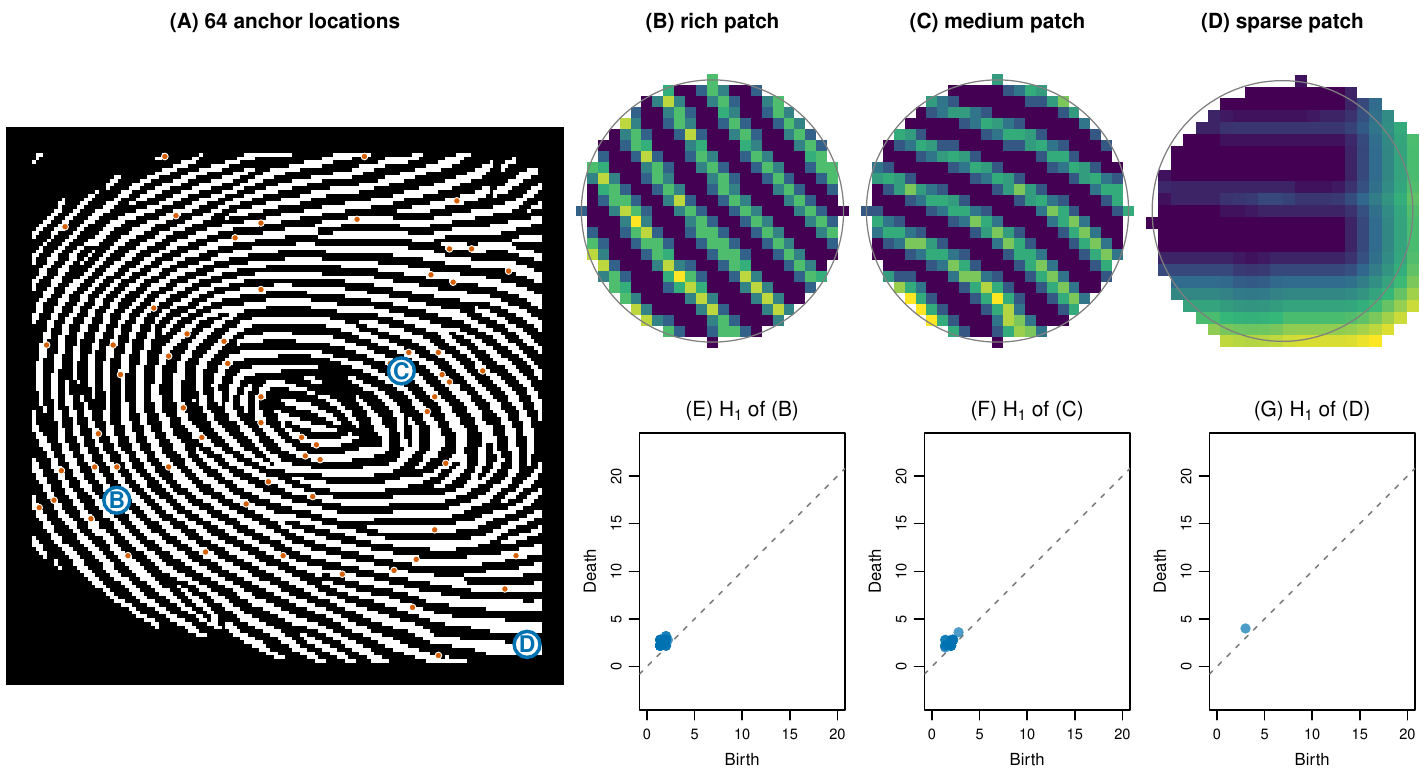}
\caption{Local patch extraction applied to an example fingerprint from FVC2000 DB1. (\emph{A})~The 64 anchor points, sampled uniformly at random from ridge pixels in the working-resolution binary image, are shown as orange dots; three anchors are highlighted and labeled (\emph{B}), (\emph{C}), and (\emph{D}). (\emph{B}--\emph{D})~The circular patches of the ridge distance transform $\mathrm{DT}_\mathrm{ridge}$ at each highlighted anchor, with the disk-mask boundary marked in grey; pixels outside the disk are assigned the maximum filtration value during the persistent-homology computation. The three patches are selected by ranking the 64 patches by total $H_1$ persistence: (\emph{B}) the highest-ranked patch, containing many short ridge segments and hence many small enclosed valley regions; (\emph{C}) the median-ranked patch; (\emph{D}) the lowest-ranked patch, drawn from a region of approximately parallel ridge flow that produces little enclosed valley space. (\emph{E}--\emph{G})~The corresponding $H_1$ persistence diagrams, showing progressively fewer features. The downstream methods use both ridge and valley channels; only ridge is shown here for brevity.}
\label{fig:patches}
\end{figure}

Local information is important because two fingerprints from the same finger should share not just similar global topology but similar \emph{spatially localized} topology. We sample 64 anchor points uniformly at random from ridge pixels in the $150 \times 150$ working-resolution binary image, providing dense spatial coverage (Fig.~\ref{fig:patches}\emph{A}). The sampling is deterministic, using an image-specific seed derived from the image identifier, ensuring reproducibility without introducing shared randomness between genuine pairs. Around each anchor, we crop a circular patch of radius 12 pixels from the \emph{full-image distance transform fields}, not recomputed per patch. This radius covers roughly 4--5 inter-ridge spacings at the working resolution, enough to capture local topological events such as bifurcations and small enclosed valleys. To prevent the patch boundary from introducing spurious topological features, we apply a disk mask: we assign pixels outside the circular disk the maximum filtration value of 20, effectively placing them at infinity in the filtration. We then compute persistent homology on this masked patch for both the ridge and valley distance transform crops, deriving local Betti curves or local persistence images as required. When a patch yields an empty persistence diagram, its Betti curve or persistence image is the zero vector. The result is a set of 64 spatially localized topological descriptors per image. Figure~\ref{fig:patches} illustrates the patch-to-patch variability in local topology that TopoLR and LPOT exploit.

\section{Verification methods: full detail}
\label{si:methods}

Table~\ref{tab:params} collects the key parameter values used in all experiments. We fixed all values through preliminary exploration before running the final experiments; we did not systematically tune any of them or conduct sensitivity analysis. This is a deliberate choice: our goal is to evaluate what topological representations contribute to fingerprint verification, not to maximize absolute performance through hyperparameter optimization.

\begin{table}[!ht]
\centering
\caption{Key parameters used in the experiments.}
\label{tab:params}
\begin{tabular}{llc}
\toprule
Component & Parameter & Value \\
\midrule
Cross-validation & Folds & 5 \\
 & Random seed & 20260110 \\
Filtration & Max scale & 20 \\
 & PH image size & $150 \times 150$ \\
 & Baseline image sizes & $50{\times}50$, $150{\times}150$ \\
Betti curves & Grid points & 50 \\
Global PI & Resolution / $\sigma$ & $20{\times}20$ / 0.5 \\
Local PI & Resolution / $\sigma$ & $30{\times}30$ / 0.25 \\
Anchors & Count / radius & 64 / 12 \\
TopoLR & Top-$K$ for pooling & 10 \\
 & $\lambda$ / batch / lr & 0.01 / 5K{+}5K / 0.1 \\
 & LR decay / max iter & $\times 0.9$ every 20 / 200 \\
 & Pair-standardizer sample & 20{,}000 training pairs \\
LPOT & $\beta$ / dustbin cost & 6 / 0.4 \\
 & Sinkhorn reg.\ / iters & 0.12 / 60 \\
 & Dustbin mass budget & 20\% \\
Fusion & Weight grid & 0 to 1 by 0.02 \\
\bottomrule
\end{tabular}
\end{table}

\subsection{Geometry baselines}

Both geometry-only baselines construct a feature vector by concatenating the ridge and valley distance transforms, normalizing by root-mean-square value, and scoring pairs by negative Euclidean distance $s(A,B) = -\|v_A - v_B\|_2$. \emph{DT Downsampled $L^2$ ($50 \times 50$)} yields a 5{,}000-dimensional vector per image; \emph{DT Downsampled $L^2$ ($150 \times 150$)} yields a 45{,}000-dimensional vector. The $50 \times 50$ baseline establishes a deliberately weak floor, while the $150 \times 150$ baseline controls for input resolution. The matched-resolution comparison still confounds the topological representation with the invariance properties it confers (main text, Discussion); any topological method that fails to beat these baselines would contribute no useful information beyond what the distance transform images already contain.

\subsection{Global topological baselines}

\emph{Global Betti Curve $L^1$.} For each image we compute the 100-dimensional global Betti vector and score a pair by $s(A,B) = -\|v_A - v_B\|_1$. We use the $L^1$ distance because the underlying Betti curve, before normalization, is a count-valued function, and the $L^1$ distance corresponds to the total discrepancy in feature counts integrated over the filtration.

\emph{Global Persistence Image $L^2$.} For each image we compute $H_1$ persistence images for ridge and valley distance transforms using a $20 \times 20$ grid with $\sigma = 0.5$, concatenated into an 800-dimensional vector, and $L^2$-normalize. The score is $s(A,B) = -\|v_A - v_B\|_2$. Comparing these two baselines provides a rough indication of whether the richer distributional information in persistence images helps discrimination, though the comparison is not fully controlled: the two methods also differ in dimensionality (100 vs.\ 800), distance metric ($L^1$ vs.\ $L^2$), and kernel bandwidth. Both methods have no learned parameters and require no training data.

\subsection{TopoLR: learned topological pair scoring}

The per-image feature vector has two parts. The \emph{global component} is the 100-dimensional global Betti vector. The \emph{local component} pools the 64 local Betti curves into summary statistics: for each distance transform, we compute the mean Betti curve across all 64 anchors, the mean curve across the top 10 anchors within that distance-transform channel, ranked by total Betti mass, and the elementwise standard deviation across all anchors, yielding $2 \times 3 \times 50 = 300$ local dimensions. We apply the same global ridge-area normalization factor to the pooled local features, using no per-patch normalization. The full per-image vector concatenates the global (100) and local (300) components for 400 dimensions.

Given two images with feature vectors $u, v \in \mathbb{R}^{400}$, we construct a symmetric pair representation invariant under interchange of the images: the elementwise absolute difference $|u - v|$, minimum $\min(u,v)$, maximum $\max(u,v)$, and product $u \odot v$, concatenated into a 1{,}600-dimensional pair feature vector. These four blocks are not all linearly independent; elementwise, $\max(u,v) - \min(u,v) = |u - v|$. The $L^2$ regularization keeps the optimization well-conditioned, and including all four blocks slightly improved performance over subsets in preliminary exploration.

We train an $L^2$-regularized logistic regression model with regularization strength $\lambda = 0.01$. Training uses mini-batch stochastic gradient descent with balanced sampling with replacement: each batch contains 5{,}000 genuine and 5{,}000 impostor pairs, drawn independently with replacement so that the same pair may appear multiple times. The learning rate starts at 0.1 and decays by a factor of 0.9 every 20 iterations, running for 200 iterations. Before training, we standardize the pair features using means and standard deviations estimated from a random sample of 20{,}000 training pairs, applied unchanged to test pairs. The output score is the logistic model output, ranging from 0 to 1. With 1{,}600 features and roughly 200{,}000 training pairs per fold, overfitting is a concern with more flexible classifiers; logistic regression with $L^2$ regularization provides a strong, interpretable linear baseline.

\subsection{LPOT: local persistence image optimal transport}

For each image, we compute a local persistence image descriptor at each of the 64 anchors. At each anchor we compute separate $H_1$ persistence images on the ridge and valley patches using a $30 \times 30$ grid with $\sigma = 0.25$, yielding 900 dimensions each, $L^2$-normalize each channel independently, and concatenate to form a 1{,}800-dimensional descriptor $d_i$. When a channel's persistence image is the zero vector because the patch yielded an empty diagram, $L^2$-normalization leaves it as the zero vector. If both anchors have an empty channel, that channel contributes zero to the descriptor cost; if only one side is empty, the channel contributes the squared norm of the other side's unit vector, exactly~1, acting as a mild penalty for the mismatch. We do not re-normalize after concatenation, so the descriptor norm is $\sqrt{2}$ when both channels are nonzero, $1$ when exactly one is nonzero, and $0$ when both are empty; we calibrate the fixed dustbin cost and geometry weight to the typical-case scale. We also record the spatial coordinates of each anchor, centering by subtracting the mean position and scaling by the image dimension of 150.

To compare two images we solve an optimal transport problem combining the descriptor cost $C^{\mathrm{desc}}_{ij} = \|d_i - d_j\|_2^2$ over the 1{,}800-dimensional descriptors and the geometry cost $C^{\mathrm{geo}}_{ij} = \|p_i - p_j\|_2^2$ over normalized positions, as $C_{ij} = C^{\mathrm{desc}}_{ij} + \beta\, C^{\mathrm{geo}}_{ij}$ with $\beta = 6$, set during preliminary exploration and held fixed across folds. We use a \emph{dustbin mechanism} that allows partial matching: we augment the cost matrix with a virtual node, with a fixed cost of 0.4 for transporting mass between any real anchor and the dustbin and zero cost for dustbin-to-dustbin transport. We fix the dustbin's mass budget at 20\% of the total and distribute the remaining 80\% equally among the 64 real anchors. Adding one dustbin node to each image's 64 anchors gives a balanced optimal transport problem on the augmented $65 \times 65$ space. At least 75\% of the real-anchor mass must still be matched real-to-real, so the dustbin handles moderate non-overlap but does not permit arbitrary partial matching. We chose the 20\% budget as a rough upper bound on the fraction of anchors expected to fall outside the overlap region in typical same-finger impression pairs, and did not tune it. We solve with the Sinkhorn algorithm using entropic regularization parameter 0.12 and exactly 60 iterations with no early stopping, ensuring exact reproducibility. The similarity score is the negative total transport cost.

\subsection{Fusion}

We normalize each method's scores via a robust z-score transformation, subtracting the median and dividing by the median absolute deviation (MAD) of the training impostor scores; MAD-based normalization is more robust to the heavy-tailed impostor distribution than mean/standard-deviation normalization. Given normalized scores $z_{\mathrm{TopoLR}}$ and $z_{\mathrm{LPOT}}$, the fusion score is $s_{\mathrm{fusion}} = w\, z_{\mathrm{TopoLR}} + (1-w)\, z_{\mathrm{LPOT}}$, with $w$ selected by grid search over $[0,1]$ in increments of 0.02 to maximize TAR@$10^{-3}$ on training data. We then apply the normalization parameters and selected weight unchanged to test data.

\section{Ablation studies}
\label{si:ablation}

To understand which components drive the performance of TopoLR and LPOT, we evaluate ablated variants of both (Table~\ref{tab:ablations}).

\begin{table}[!ht]
\centering
\caption{Ablation results for TopoLR and LPOT (mean $\pm$ SD across 5 folds). EER and TAR values are percentages.}
\label{tab:ablations}
\begin{tabular}{lccc}
\toprule
Variant & AUC & EER & TAR@$10^{-3}$ \\
\midrule
\multicolumn{4}{l}{\emph{TopoLR ablations}} \\
\quad Global+Local (full) & .906$\pm$.011 & 17.4$\pm$1.6 & 4.9$\pm$4.2 \\
\quad Global only & .904$\pm$.013 & 17.7$\pm$1.6 & 4.0$\pm$3.3 \\
\quad Local only & .886$\pm$.015 & 19.3$\pm$1.8 & 3.4$\pm$2.8 \\
\quad No Top-$K$ & .907$\pm$.012 & 17.2$\pm$1.5 & 4.2$\pm$3.2 \\
\midrule
\multicolumn{4}{l}{\emph{LPOT ablations}} \\
\quad Full & .836$\pm$.008 & 25.2$\pm$1.2 & 11.2$\pm$4.8 \\
\quad No geometry & .783$\pm$.014 & 29.5$\pm$1.7 & 2.3$\pm$1.6 \\
\quad No dustbin & .827$\pm$.011 & 25.7$\pm$1.7 & 7.5$\pm$3.4 \\
\quad Ridge only & .813$\pm$.009 & 27.1$\pm$1.5 & 8.9$\pm$5.1 \\
\quad Valley only & .802$\pm$.009 & 27.6$\pm$0.6 & 4.9$\pm$1.9 \\
\bottomrule
\end{tabular}
\end{table}

\emph{TopoLR ablations.} Removing local features causes only a small AUC drop, from 0.906 to 0.904, well within the fold-to-fold standard deviations, so global Betti features carry most of the discriminative information the logistic model exploits, though the difference is too small relative to variance to be conclusive. Using local features alone achieves AUC 0.886, below the global-only variant by 0.018, a larger gap but still comparable to the standard deviations. The top-$K$ pooling mechanism has essentially no effect (AUC 0.907 without top-$K$ vs.\ 0.906 with it), meaning the simpler scheme of pooling only the mean and standard deviation across anchors would suffice. After pooling, the model operates on aggregate statistics over anchors rather than individual patch identities, so any effective feature selection acts on filtration-bin dimensions of the pooled Betti vector, not on specific spatial patches.

\emph{LPOT ablations.} The geometry cost is critical: removing it drops AUC from 0.836 to 0.783 and TAR@$10^{-3}$ from 11.2\% to 2.3\%. The AUC drop of 0.053 is large relative to the fold-to-fold variability, and the TAR collapse is consistent across all TAR thresholds we examined, not only the TAR@$10^{-3}$ reported in the table. Without spatial coherence, the optimal transport solver can match topologically similar but spatially inconsistent regions, for instance aligning a loop on the left side of one fingerprint to a similar loop on the right side of another. The dustbin mechanism contributes meaningfully but less dramatically: removing it reduces TAR@$10^{-3}$ from 11.2\% to 7.5\%. The ability to leave some anchors unmatched helps LPOT handle noise, partial overlap, and regions where one impression captured skin that the other did not. Within LPOT, the ridge channel carries more low-FAR signal than the valley channel: ridge-only achieves TAR@$10^{-3}$ of 8.9\% versus 4.9\% for valley-only. This is consistent with ridge structure being the primary carrier of fingerprint identity, since pattern type, minutiae, and ridge flow are all properties of the ridges, while valleys are defined negatively as the complement. Nevertheless, combining both channels performs best within LPOT, showing that the valley channel contributes additional discriminative information in this matching model.

\section{Runtime}
\label{si:runtime}

Table~\ref{tab:runtime} reports wall-clock time per fold, excluding the one-time feature precomputation that we run once per image regardless of how many pairs we score. For methods that require no learning, the reported time covers pair scoring only. For TopoLR, it includes pair-feature construction, standardization, logistic regression training, and scoring. For Fusion, it is the marginal cost of z-normalization and linear combination given that TopoLR and LPOT scores already exist.

\begin{table}[!ht]
\centering
\caption{Runtime per fold in seconds (mean $\pm$ SD across 5 folds), excluding feature precomputation.}
\label{tab:runtime}
\begin{tabular}{lc}
\toprule
Method & Seconds \\
\midrule
Betti $L^1$ & 0.1$\pm$0.1 \\
PI $L^2$ & 1.1$\pm$0.2 \\
DT Down.\ $L^2$ ($50^2$) & 5.2$\pm$0.6 \\
DT Down.\ $L^2$ ($150^2$) & 32.4$\pm$1.2 \\
TopoLR & 82.0$\pm$3.6 \\
LPOT & 143.1$\pm$1.8 \\
Fusion & 0.2$\pm$0.0 \\
\bottomrule
\end{tabular}
\end{table}

The global no-learning methods are fastest, completing in about one second or less per fold. The geometry baselines are slower because of their large feature vectors (5{,}000 and 45{,}000 dimensions). TopoLR requires about a minute and a half per fold, dominated by constructing 1{,}600-dimensional pair features and the logistic regression loop. LPOT is the most expensive at roughly two and a half minutes per fold, reflecting the cost of solving an optimal transport problem for each pair. Fusion adds negligible overhead. We measured all timings on an Apple Mac Studio with an M2 Ultra processor and 192\,GB of RAM. Feature precomputation, computing distance transforms, running persistent homology via GUDHI, and extracting local patches and descriptors for all 800 images, takes approximately 30 seconds using 22 parallel workers and runs only once.

\section{Dimension-matched comparison of global summaries}
\label{si:pca}

Betti curves and persistence images achieve nearly identical AUC (0.847) but differ substantially at low FAR. Because the two representations differ simultaneously in dimensionality (100 vs.\ 800), distance metric ($L^1$ vs.\ $L^2$), and kernel parameters, the gap could in principle reflect any of these. To partially disentangle them, we projected each representation onto its own 50-dimensional basis estimated by principal component analysis on training-fold features and scored test pairs with the same $L^2$ metric. Under this matched-dimension comparison (Table~\ref{tab:pca}), the AUC values change little relative to fold-to-fold variability and the gap at TAR@$10^{-3}$ persists. This suggests that the low-FAR difference between the Betti and persistence-image baselines is not primarily an artifact of dimensionality or metric, but reflects the representations themselves: at strict thresholds, the count-based Betti summary appears more resistant to noise than the smoothed joint birth--persistence distribution captured by persistence images, possibly because the higher-dimensional, smoother persistence image is more susceptible to noise at the extreme tail of the impostor distribution.

\begin{table}[!ht]
\centering
\caption{Dimension-matched comparison: each global representation projected onto its own 50-dimensional PCA basis, estimated on training-fold features, and scored with the $L^2$ metric (mean $\pm$ SD across 5 folds). EER and TAR are percentages.}
\label{tab:pca}
\begin{tabular}{lccc}
\toprule
Variant & AUC & EER & TAR@$10^{-3}$ \\
\midrule
Betti (PCA-50, $L^2$) & .854$\pm$.014 & 23.0$\pm$1.4 & 5.5$\pm$1.3 \\
PI (PCA-50, $L^2$) & .848$\pm$.015 & 23.0$\pm$1.7 & 3.0$\pm$0.5 \\
\bottomrule
\end{tabular}
\end{table}

\clearpage
\bibliographystyle{unsrtnat}
\bibliography{references}
\end{document}